%% file: acl2021.tex
\documentclass[11pt,a4paper]{article}
\usepackage[hyperref]{acl2021}
\usepackage{times}
\usepackage{latexsym}
\usepackage{balance}
\usepackage{subfigure}
\usepackage{amsmath}
\usepackage{amssymb}
\usepackage{caption}
\usepackage{graphicx}
\usepackage{url}
\usepackage{hyperref}
\usepackage{multicol}
\usepackage{multirow}
\usepackage{booktabs}
\usepackage{color}
\usepackage{array}
\usepackage{mathtools} 
\usepackage{pythonhighlight} 
\usepackage{listings}

\usepackage{microtype}

\aclfinalcopy 

\title{SemEval-2023 Task 4: Fine-tuning vs Prompting,\\
Can Language Models Understand Human Values?}

\author{Pingwei SUN\\ 
        BDT program, HKUST\\ \texttt{psunah@connect.ust.hk} \\
   }
  
\date{}

\begin{document}
\maketitle


\begin{abstract}
Accurately handling the underlying support values in sentences is crucial for understanding the speaker's tendencies, yet it poses a challenging task in natural language understanding (NLU). In this article, we explore the potential of fine-tuning and prompt tuning in this downstream task, using the \emph{\href{https://touche.webis.de/semeval23/touche23-web/index.html}{\textcolor{black}{Human Value Detection 2023}}}. Additionally, we attempt to validate whether models can effectively solve the problem based on the knowledge acquired during the pre-training stage. Simultaneously, our interest lies in the capabilities of large language models (LLMs) aligned with RLHF in this task, and some preliminary attempts are presented. 
\end{abstract}

\input{paragraph/introduction}
\input{paragraph/relatedwork}

\input{paragraph/method}
\input{paragraph/experiment}
\input{paragraph/analysis}
\input{paragraph/conclusion}

\bibliographystyle{acl_natbib}
\bibliography{acl2021}


\end{document}

%% file: paragraph/introduction.tex
\section{Introduction}\label{Sec1}
The persuasiveness of arguments is heavily influenced by individuals' values, and variations in the priority assigned to these values can lead to disputes, especially between different cultures. To address this, computational linguistics leverages human values to categorize and evaluate arguments.

Schwartz's value categorization\citep{schwartz1994there} serves as a common framework. However, existing datasets have limitations, such as small sizes and cultural biases. To overcome these issues, the competition holder proposes an extension, Touché23-ValueEval~\citep{mirzakhmedova2023touche23valueeval}, containing 9324 diverse arguments from various sources and cultures. This dataset aims to facilitate the development of Pre-trained Language Models (PLM) for the automatic identification of human values in persuasive communication, supporting a broad array of applications. The task attracted many researchers to work on it and various methods were proposed to solve the problem. 

However, from the leaderboard, this task still requires more in-depth research, with the top performer only achieving an average F1 score of 0.56. Most participating teams tried to solve it as a classification task and employed techniques such as model ensembling and expanding the dataset to improve scores. Taking into account computational resources and the rise of LLMs, our project will focus on the following questions based on the dataset:
\begin{itemize}
    \item\textbf{Q1:} Does prompt-tuning work compared with fine-tuning on the complex downstream task?
    \item\textbf{Q2:} Can PLMs handle human values powered by knowledge from the pre-training stage?
    \item\textbf{Q3:} How do LLMs perform on the task after aligning with human preference?
\end{itemize}

%% file: paragraph/relatedwork.tex
\section{Related Work}

\begin{table*}[ht]
\begin{tabular}{lcp{10cm}}
\toprule[1.5pt]
Feature      & Value Type & Example                                                                       \\ \hline
Argument ID  & Str & "A01002"                                                                             \\
Conclusion   & Str & "We should ban fast food"                                                            \\
Stance       & Str & {[}"supporting", "against"{]}                                                        \\
Premise      & Str & "Fast food should be banned because it is really bad ..." \\
Value Labels & Int & [0, 1]\newline
                     Value categories: "Achievement", "Stimulation", etc. \\
Value Description &
  Str & "Self-direction: thought. It is good to have own ideas and interests. \newline
  Contained values and associated arguments (examples): ..."\\
\bottomrule[1.5pt]
\end{tabular}
\caption{The dataset used in the experiment with the corresponding types and examples. \textbf{Stance} is mapped to single words to simplify the prompting template. Square brackets represent single selections from among. The \textbf{Value Description} are defined in \citealp{schwartz1994there}.}
\label{Tab1}
\end{table*}

\textbf{Human Value Detection 2023.}
Among the submissions, most teams applied transformers-based models as their backbone and treated it as a classification task.

Team Adam Smith~\citep{schroter:2023} employs an ensemble strategy, including DeBERTa and RoBERTa. They trained these models for loss minimization or F1-score maximization on three folds each RoBERTa model was pretrained on the IBM-ArgQ-Rank30KArgs dataset. Ensembling involved averaging predictions with an optimized decision threshold. They experimented with a stacked meta-classifier based on logistic regression. 

Team John Arthur~\citep{balikas:2023} fine-tuned a DeBERTa model on the task's data, using a concatenated representation of stance, premise, and conclusion. They found that separate token symbols for stance improved classification. The model was trained to minimize binary cross-entropy loss, and the team observed performance benefits with more training data. 

Team PAI~\citep{ma:2023} combined models using various input datasets. They applied weight voting based on F1 score for ensembling and explored different loss functions, ultimately favoring a class-balanced loss. Classification thresholds were tested, but no performance improvement was achieved.

\textbf{Fine-tuning.}
Since the release of BERT~\citep{devlin2018bert}, the approach of large-scale pre-training followed by fine-tuning on downstream tasks has been widely explored and applied to various NLP subtasks. 

Through the integration of a range of training techniques, this method has achieved impressive results. The bi-encoder structure employed during pre-training inherently endows it with the ability to effectively extract semantic information from sequences. However, it performs poorly when facing few-shot and zero-shot scenarios.

\textbf{Prompting.}
Recently, researchers have been attempting to transform NLP tasks into a seq-to-seq format using prompts. T5~\citep{raffel2023exploring} marked the first successful endeavor in this direction. Such an approach leverages the knowledge acquired by language models during the pre-training phase rather than treating them as feature extractors. Consequently, it exhibits excellent performance on few-shot and zero-shot tasks, with the added benefit of reduced fine-tuning costs. 

GPT, with its decoder structure, demonstrates enhanced capabilities in language modeling. The team Hitachi~\cite{tsunokake:2023} has used BART, T5, and GPT3, feeding the data to them in question-answering format.

%% file: paragraph/method.tex
\section{Methods}
\begin{figure*}[htb]
\centering 
\includegraphics[width=1.0\linewidth]{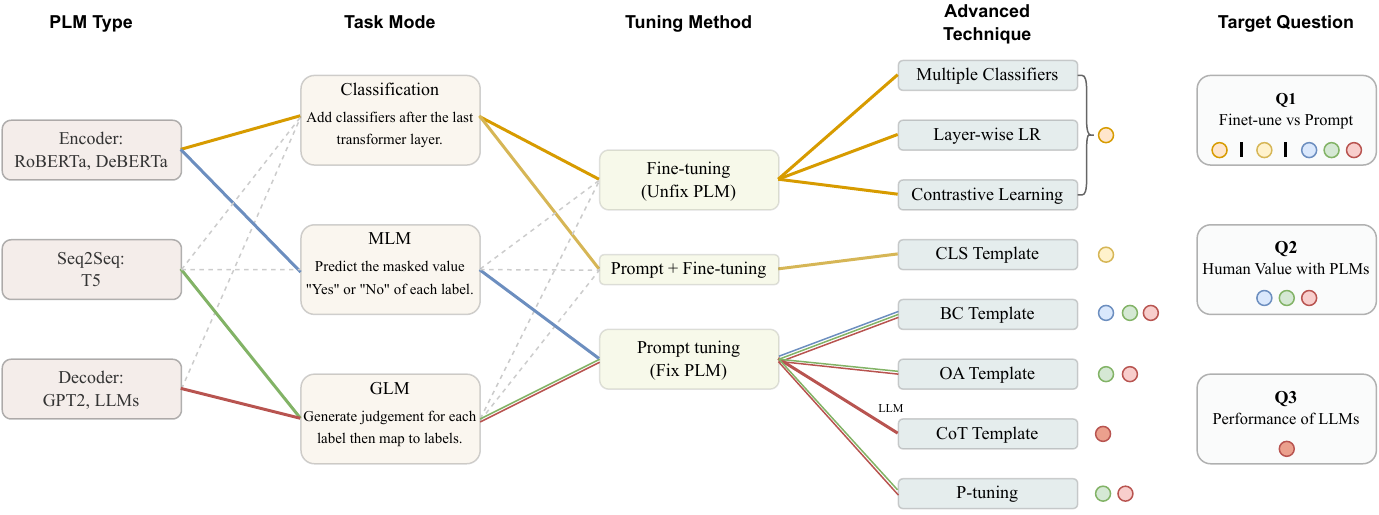}
    \caption{Illustration of the workflow of our experiments. Tracks marked by different colors stand for combinations of models, tasks, and methods, details of which are told in § \ref{Sec4}. The dashed lines mean methods are theoretically available but we do not consider them in the project because of poor performance or being computationally demanding. The colorful dots are experiment results and indicate the analysis for the questions proposed in § \ref{Sec1}.}
    \label{expflow}
\end{figure*}

To answer the questions proposed in § \ref{Sec1}, we propose the following techniques, to our best knowledge, to measure the potential of models on the complex task of detecting human values. Figure \ref{expflow} illustrates our workflow in the project.

\subsection{Data Utilization}
The data file contains various annotations and supplementary datasets from platforms like Zhihu and The New York Times. Experiments are conducted on the main dataset, and due to the unavailability of test labels, all results in this paper reflect the model's performance on the validation set. 

Before commencing the experiments, the data files are reorganized, and certain features are preprocessed, as shown in Table \ref{Tab1}.

\subsection{Fine-tuning as a Classification Task}
\textbf{Classifier.}
To solve the multi-label classification task, there are two methods to choose from. On the one hand, it is sensible to use a single classifier, apply the sigmoid function to each $x_i$ in the $output=\left \{x_1, x_2, ..., x_c  \right \}$, and then calculate the Binary Cross Entropy Loss as follows. 
\begin{equation}
\sum_{i=1}^{C} y_{i} \cdot \log \sigma (x_{i}) + (1 - y_{i}) \cdot \log (1 - \sigma (x_{i}))
\end{equation}

On the other hand, it also works by setting up the same number of classifiers as the number of labels, each classifier to handle one category.

\textbf{Hidden States.}
Serving as an encoding module, BERT can provide various levels of features of the sequence. It is commonly agreed that the deeper layers always output high-level semantic information~\citep{sun2019fine} and the first token [CLS] can be used for sequence classification.

\textbf{Avoid Forgetting.}
Since the model has learned solid knowledge during pre-training, we need to find an appropriate way to arrange the learning rate for fine-tuning. The lower layers handle more general information so they should be updated with relatively lower learning rates. We use the decay factor to achieve layer-wise learning rates.

\subsection{Contrastive Learning}
This project is a complex Natural Language Understanding task, so obtaining a good embedding representation of the sequence is crucial for the subsequent training of the classifier.

Inspired by the SimCSE~\citep{gao2021simcse}, Contrastive Learning (CL) loss is applied to optimize the embedding states of sequences. Since the target is multi-label classification, the positive and negative samples can not be defined directly inside a mini-batch. Consequently, we compute the CL loss according to the following formulas:
\begin{equation}
\begin{split}
    w_{ij} =& \frac{y_{i}^{T}\cdot y_{j}}{\sum_{k=1}^{B} y_{i}^T \cdot y_k + \varepsilon} \\
    {\large CL\ \ell} _{i} = -&\log \frac{\sum_{j=1}^{B} w_{ij}\cdot e^{\frac{sim(x_i, x_j)}{\tau'}} }{\sum_{j=1}^{B} e^{\frac{sim(x_i, x_j)}{\tau'}}} 
\end{split}
\end{equation}
where $y$ is the label vector, $x$ is the embedding representation and $B$ is the batch size. The $\tau '$ stands for the temperature hyper-parameter of CL, which is a scaling factor of similarity scores.

\subsection{Prompt Tuning Techniques}
\begin{figure*}[htb]
\centering 
\includegraphics[width=1.0\linewidth]{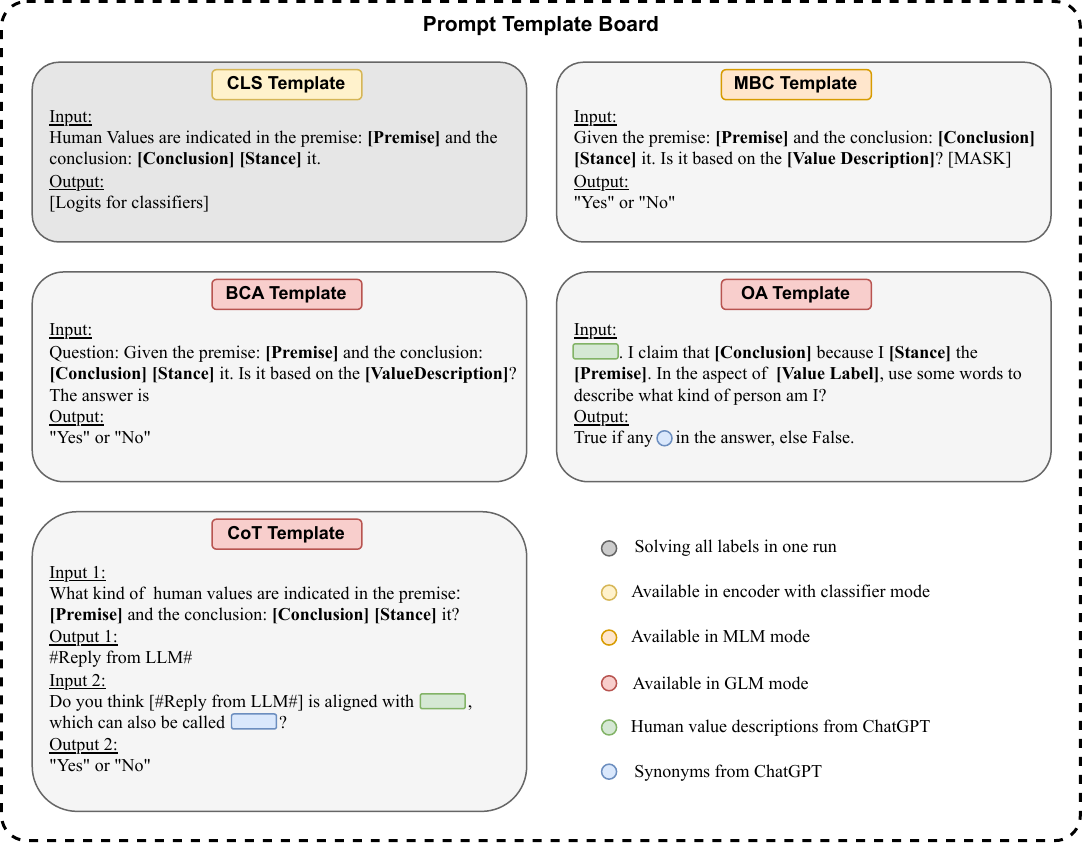}
    \caption{Prompt templates for different task processing modes, including classification(CLS), masked binary choice(MBC), binary choice answering(BCA), open answering(OA), and Chain-of-Thought(CoT). The bolded content in brackets represents the features in the dataset, shown in Table \ref{Tab1}.}
    \label{promptboard}
\end{figure*}

When doing prompt tuning, the model should be transferred to other modes to fit the template's requirements, shown in Figure \ref{promptboard}. The details of template construction are introduced below.

\textbf{Binary choice template.}
It is easy to come up with the idea to transfer the multi-label classification task to several binary choice questions. It can be solved either with a Masked Language Model(MLM) by filling the blank or a Generative Language Model(GLM) by generation.

\textbf{Open answering template with knowledgeable verbalizer.}
When our templates become more complex and open, it often allows for better stimulation of internal model knowledge. At this point, we need a verbalizer to map the output results to labels, and the choice of verbalizer significantly impacts the outcomes~\citep{hu-etal-2022-knowledgeable}.

Notably, the dataset includes descriptions and examples of human values. However, they were artificially defined by linguists two decades ago. To enhance the dataset's generalization, we introduce LLM to rewrite the descriptions and provide synonyms, which will serve as a knowledgeable verbalizer to map the output results to labels of human values.

\textbf{Chain-of-Thought template.}
Language models, particularly LLMs, have demonstrated remarkable reasoning capabilities, as evidenced by studies such as~\citep{kojima2022large}. Despite the challenge of directly comprehending whether specific human values are implied in triples, models can acquire relevant knowledge through a systematic prompting process. This step-by-step approach allows LLMs to navigate and extract pertinent information, ultimately enabling them to effectively address the intricacies of classification problems.

In light of this, we have incorporated templates in the form of CoT. These templates serve as structured prompts that guide the model in interpreting and understanding the nuances of human values within extra knowledge.

%% file: paragraph/experiment.tex
\section{Experiments}\label{Sec4}
 The following experiments are conducted to answer questions mentioned in § \ref{Sec1}. Since we aim to explore fine-tuning methods and have a limited computational budget, tricks such as model ensembling to improve scores are not applied.

\subsection{Exp-\uppercase\expandafter{\romannumeral1}: Fine-tuning}\label{Sec41}
Based on the previous team's papers, we choose RoBERTa and DeBERTa followed by a classifier for fine-tuning experiments. The Premise, Conclusion, and Stance features are directly fed to the model. Settings of hyperparameters for training are listed in Table \ref{hyperparam}. 

The performance of different models and classifiers is presented in Table \ref{tab:baseline}. RoBERTa with multiple classifiers is selected after the preliminary fine-tuning experiment.

Then Contrastive Learning is introduced for further improvement of the performance. In this part, we treat it as either a pre-training task or an extra loss item at the fine-tuning stage. The second strategy performs slightly better, and both of them make progress on the macro F1 score.

\begin{table}[!h]
\centering
\begin{tabular}{lc}
\toprule[1.5pt]
\textbf{Parameters}       & \textbf{Value}    \\ \hline
\multicolumn{2}{c}{\emph{Trainer}}            \\
Batch size      & 8                           \\
Epochs          & 3.0                         \\
Lr scheduler    & cosine                      \\
Warmup ratio    & 0.1                         \\
Lr decay        & 0.97                        \\
\multicolumn{2}{c}{\emph{Optimizer}}          \\
Optimzier        & Adamw                      \\
Learning rate    & 2e-5                       \\
Trainable param  & last 8 layers + classifier(s) \\
\bottomrule[1.5pt]
\end{tabular}%
\caption{We try various combinations of the hyperparameters and compare validation scores after training. The above is an optimal choice among them.}
\label{hyperparam}
\end{table}

\begin{table}[hbt]
\centering
\begin{tabular}{lc}
\toprule[1.5pt]
\textbf{Model} & \textbf{Macro F1 score} \\ \hline
$RoBERTa_{Large}$        &                       \\
\quad             w/ SH  &  .481                 \\
\quad             w/ MH  &  \textbf{.507}        \\
$DeBERTa_{Large}$                                \\
\quad             w/ SH  &  .477                 \\
\quad             w/ MH  &  .493                 \\ \hline
$RoBERTa_{Large}$ w/ MH                          \\
\quad CL pre-train       &  .518                 \\
\quad CL fine-tune       &  \textbf{.522}        \\
\bottomrule[1.5pt]
\end{tabular}%
\caption{Performance of finetuned encoding models with classifiers on the validation set. SH means single classification head, while MH means multiple heads.}
\label{tab:baseline}
\end{table}

\input{paragraph/prompttable}

\subsection{Exp-\uppercase\expandafter{\romannumeral2}: Prompt Tuning}
The previously mentioned prompt templates are sequentially tested in this section in conjunction with models of different parameter sizes and structures. The training strategies are inspired by \citealp{liu2021pretrain} and implementations are based on OpenPrompt~\citep{ding2021openprompt}.

To be specific, the CLS template is treated as a hard prompt of which parameters are frozen and fed to encoding models with trainable multiple classifiers. Its trainable parameters are consistent with the previous fine-tuning settings. 

Templates in the forms of mask filling and question answering are also proposed. At this time, we set PLMs fixed and allow prompts’ parameters to be updated. Extra knowledge in the OA template is processed in advance, which is manually selected from feedback from ChatGPT.

\subsection{Exp-\uppercase\expandafter{\romannumeral3}: Works on LLMs}
The capabilities of the LLMs on this complex NLU task are also what we plan to explore. Experiments are conducted with a CoT template which is designed to stimulate the logical inference ability and knowledge inside LLMs. 

Considering the computational budget, no gradient updating process is included in this template. Instead, the scores are obtained by taking 5\% of the validation dataset for local inference (Llama7B) or API calls (GPT series) and computing F1 scores on predicted results, which is shown in Table \ref{promptexp}.

%% file: paragraph/prompttable.tex
\begin{table*}[htb]
\centering
\resizebox{\textwidth}{!}{%

\begin{tabular}{llllllllllllllllllllll}

\toprule[1.5pt]
\textbf{Method\ /\ F1 score} &
  \textbf{{All}} &
  \textbf{\rotatebox{90}{Self-direction: thought}} &
  \textbf{\rotatebox{90}{Self-direction: action}} &
  \textbf{\rotatebox{90}{Stimulation}} &
  \textbf{\rotatebox{90}{Hedonism}} &
  \textbf{\rotatebox{90}{Achievement}} &
  \textbf{\rotatebox{90}{Power: dominance}} &
  \textbf{\rotatebox{90}{Power: resources}} &
  \textbf{\rotatebox{90}{Face}} &
  \textbf{\rotatebox{90}{Security: personal}} &
  \textbf{\rotatebox{90}{Security: societal}} & 
  \textbf{\rotatebox{90}{Tradition}} &
  \textbf{\rotatebox{90}{Conformity: rules}} &
  \textbf{\rotatebox{90}{Conformity: interpersonal}} &
  \textbf{\rotatebox{90}{Humility}} &
  \textbf{\rotatebox{90}{Benevolence: caring}} &
  \textbf{\rotatebox{90}{Benevolence: dependability}} &
  \textbf{\rotatebox{90}{Universalism: concern}} &
  \textbf{\rotatebox{90}{Universalism: nature}} &
  \textbf{\rotatebox{90}{Universalism: tolerance}} &
  \textbf{\rotatebox{90}{Universalism: objectivity}} \\ \hline
\emph{Classification}      &  &  &  &  &  &  &  &  &  &  &  &  &  &  &  &  &  &  &  &  &  \\
\quad RoBERTa w/ MH        & .51 & - & - & - & - & - & - & - & - & - & - & - & - & - & - & - & - & - & - & - & - \\
\quad RoBERTa-CLS          & \textbf{.54} & - & - & - & - & - & - & - & - & - & - & - & - & - & - & - & - & - & - & - & - \\
\quad DeBERTa-CLS          & .52 & - & - & - & - & - & - & - & - & - & - & - & - & - & - & - & - & - & - & - & - \\
\emph{MLM}                 &  &  &  &  &  &  &  &  &  &  &  &  &  &  &  &  &  &  &  &  &  \\
\quad RoBERTa-MBC          & .47 & \underline{.52} & .63 & .22 & .31 & .58 & .33 & \textbf{.56} & .30 & .69 & .62 & \underline{.58} & \underline{.55} & .27 & .16 & \underline{.50} & .32 & .66 & .75 & .39 & .43 \\
\quad DeBERTa-MBC          & - & - & - & - & - & - & - & - & - & - & - & - & - & - & - & - & - & - & - & - & - \\
\emph{GLM}                 &  &  &  &  &  &  &  &  &  &  &  &  &  &  &  &  &  &  &  &  &  \\
\quad T5$^{\dag}$-BCA      & .45 & .49 & .59 & \underline{.23} & .30 & .58 & .37 & .50 & .25 & .70 & \underline{.61} & .44 & .46 & .25 & \underline{.21} & .48 & .28 & .68 & .73 & .33 & .50 \\
\quad T5$^{\dag}$-OA       & .45 & .49 & .58 & \textbf{.25} & .31 & .58 & .32 & .49 & .27 & .73 & .60 & .49 & .45 & .25 & .19 & .47 & .28 & .66 & .74 & \underline{.35} & .52 \\
\quad T5-BCA               & \underline{.52} & \textbf{.57} & \textbf{.70} & .18 & \textbf{.44} & \textbf{.65} & \underline{.41} & \underline{.54} & \textbf{.33} & \textbf{.78} & \textbf{.68} & \textbf{.67} & \textbf{.59} & .30 & .13 & \underline{.50} & .35 & \textbf{.76} & \textbf{.85} & \underline{.49} & \textbf{.56} \\
\quad T5-OA                & .50 & .51 & .69 & .13 & .27 & \underline{.61} & \textbf{.44} & .49 & .30 & \underline{.74} & .62 & \underline{.58} & .55 & \textbf{.36} & .17 & \textbf{.52} & \textbf{.41} & \underline{.73} & .80 & .45 & \underline{.55} \\
\quad GPT2$^{\dag}$-BCA    & .42 & .41 & .57 & .15 & .21 & .55 & .26 & .43 & .23 & .66 & .53 & .47 & .49 & .31 & .20 & .42 & .28 & .64 & .72 & .36 & .53 \\
\quad GPT2$^{\dag}$-OA     & .44 & .45 & .62 & .11 & \underline{.42} & .47 & .31 & .41 & .23 & .71 & .56 & .44 & .51 & .33 & .20 & .45 & .30 & .67 & .77 & .39 & .52 \\
\quad GPT2-BCA             & .46 & .44 & .63 & .19 & .23 & .58 & .38 & .47 & .25 & .69 & .56 & .52 & .54 & \underline{.34} & \textbf{.23} & .43 & .28 & .68 & .79 & .32 & \textbf{.56} \\
\quad GPT2-OA              & .48 & .47 & \underline{.65} & \textbf{.25} & .29 & .56 & .37 & .51 & \underline{.31} & .72 & .60 & .51 & \underline{.55} & .27 & .19 & \underline{.50} & .33 & .67 & \underline{.81} & \textbf{.52} & .54 \\ \hline
\emph{LLM-CoT}             &  &  &  &  &  &  &  &  &  &  &  &  &  &  &  &  &  &  &  &  &  \\
\quad Llama-7B             & .49 & .55 & .66 & .20 & .39 & .64 & .35 & .48 & .26 & .77 & .63 & .54 & .53 & .35 & .20 & .53 & .38 & .71 & .80 & .34 & .42 \\
\quad ChatGPT              & .53 & .52 & .71 & .26 & .33 & .61 & .48 & .52 & .38 & .75 & .65 & .60 & .56 & .47 & .16 & .55 & .42 & .72 & .83 & .42 & .57 \\
\bottomrule[1.5pt]
                  
\end{tabular}%
}
\caption{F1 score of various prompt tuning methods. T5 and GPT2 with $^{\dag}$ stand for the base and medium versions respectively and the others are of their large versions except for LLMs. There is something wrong with \href{https://github.com/huggingface/transformers/issues/22790}{DeBERTaForMaskedLM in huggingface}, so DeBERTa-MBC is of no reference value.}
\label{promptexp}
\end{table*}

%% file: paragraph/analysis.tex
\section{Analysis}
\subsection{Effectiveness of Fine-tuning and Input Description}
We achieved a relatively competitive result compared with last year's leaderboard. At the same time, we also validated the contribution of contrastive learning to obtaining more discriminative embedding representations.

Furthermore, we observed that, in handling NLU tasks with complex inputs, providing reasonable descriptions for complex input examples enhances model performance compared to directly concatenating and inputting individual components.

\subsection{Performance of Prompt Tuning}
Finding versions of different models with the same parameter size is a challenge, to ensure the validity of the results, we attempted to compensate by the trainable parameters during training.

From the table, it is evident that on models with a size less than 500M, fine-tuning outperforms prompt tuning, which is reasonable being given more trainable parameters. However, when the model size approaches 1B (two to three times that of the formers), prompt tuning shows comparable performance to fine-tuning. Yet, during training, we observed a larger variance in the performance of prompt tuning with changes in the initialization methods of templates.

Does it mean fine-tuning is always a better choice in this range of parameter size (\textless 1B)? Our two additional experiments explain this in more detail. The results of NLI and few-shot are listed in Table \ref{extraexp}. 

We built a simple NLI task using the Premise Conclusion and Stance features in the dataset and tuned some models using the same method as before. As can be seen, there is no significant gap between fine-tuning and prompt tuning in this simpler task, even with fewer trainable parameters, indicating that with the increase in task complexity, prompt tuning requires a larger base model, meaning better language modeling abilities and more knowledge to achieve performance comparable to the encoder + classifier paradigm.

When faced with few-shot datasets, prompt tuning shows obvious advantages. In contrast, fine-tuning seems failed to solve the task.

\subsection{Prompting LLMs to Detect Human Values}
In the aspect of LLMs, even without any fine-tuning, a well-designed questioning style (prompts) can enable them to perform well on the task. 

However, this improvement is more uncertain than the previously mentioned methods (as can be seen from the variation in scores across categories) and relies on an experienced questioner. Failures in a few categories may be attributed to inconsistencies between external knowledge from ChatGPT and task definitions.

\begin{table}[hbt]
\centering
\begin{tabular}{lcc}
\toprule[1.5pt]
\textbf{Models / Task} & \textbf{NLI} & \textbf{5-shot HVD} \\ \hline
RoBERTa                &      .872    &        -            \\
T5                     &      .866    &        .413         \\
GPT2                   &      .849    &        .408         \\ 
\bottomrule[1.5pt]
\end{tabular}%
\caption{Results of NLI and few-shot Human Value Detection. The model sizes are close and all below 500M. NLI is evaluated by accuracy while HVD still uses the F1 score.}
\label{extraexp}
\end{table}

%% file: paragraph/conclusion.tex
\section{Conclusion}
\subsection{Fine-tuning vs Prompt Tuning}
This is a problem that involves multiple variables as we mentioned in the analysis section. Each approach has its own strengths and weaknesses. 

Briefly, prompt tuning can achieve and sometimes surpass the performance of direct fine-tuning when the model size is large enough. It exhibits better generalization and enhanced transferability. However, direct fine-tuning, as a more straightforward approach, is worth considering when training samples are abundant, as it allows for stable performance leveraging a smaller model, which is particularly crucial in industrial applications.

\subsection{Human values detection}
\textbf{PLMs of medium size:} From the F1 scores, it is evident that the model requires further improvement. However, the accuracy metric in the experiments reflects the model's detection capability is viable. The main direction for improvement lies in reducing false positives.\\
\textbf{Chat-LLMs:} Despite evaluating only a small subset of data, LLMs have demonstrated commendable performance in understanding human values.

\subsection{Limitations}
When comparing the performance of fine-tuning and prompt tuning, the inability to control model parameter sizes and trainable parameters leads to a decrease in the persuasiveness of the results. 

However, such conclusions remain meaningful for practical applications because, in the face of various trade-off conditions, considerations across all aspects are often needed rather than an itemized comparison of pros and cons.